\pdfoutput=1

\documentclass[11pt]{article}

\usepackage[]{acl}

\usepackage{times}
\usepackage{latexsym}

\usepackage[T1]{fontenc}

\usepackage[utf8]{inputenc}

\usepackage{microtype}

\usepackage{amsmath,amssymb,amsfonts}
\usepackage{url,enumitem}
\usepackage{booktabs}
\usepackage{xspace}
\usepackage{graphicx}
\usepackage{subcaption}
\usepackage{comment}
\usepackage[normalem]{ulem}
\newcommand{\Sref}[1]{\S\ref{#1}}
\DeclareMathOperator{\argmin}{argmin}

\newcommand{\methodname}[1]{ORCA}

\title{ORCA: Interpreting Prompted Language Models via Locating\\Supporting Data Evidence in the Ocean of Pretraining Data}

\author{
  Xiaochuang Han \quad \quad \quad \quad Yulia Tsvetkov\\
  Paul G.~Allen School of Computer Science \& Engineering, University of Washington \\
        {\tt \{xhan77, yuliats\}@cs.washington.edu}
}

\begin{document}
\maketitle

\begin{abstract}

Large pretrained language models have been performing increasingly well in a variety of downstream tasks via prompting. 
However, it remains unclear \emph{from where} the model learns the task-specific knowledge, especially in a zero-shot setup. 
In this work, we want to find evidence of the model's task-specific competence from pretraining 
and are specifically interested in locating a very small subset of pretraining data that directly supports the model in the task. 
We call such a subset \emph{supporting data evidence} and propose a novel method \methodname{} to effectively identify it, by iteratively using gradient information related to the downstream task. 
This supporting data evidence offers interesting insights about the prompted language models: in the tasks of sentiment analysis and textual entailment, BERT shows a substantial reliance on BookCorpus, the smaller corpus of BERT's two pretraining corpora, as well as on pretraining examples that mask out synonyms to the task verbalizers.\footnote{Code will be available at \url{https://github.com/xhan77/orca}.}

\end{abstract}

\section{Introduction}

Contemporary large language models (LLMs) are trained on massive text corpora from the web, referred to as the pretraining data \citep[e.g.,][]{Devlin2019BERTPO,Raffel2020ExploringTL}. Due to their volume, these data typically cannot be inspected manually and are prone to spelling or logic errors, biases, irrelevance to downstream tasks, and other artifacts \citep{Bender2021OnTD}. Yet, LLMs pretrained with such noisy data attain good performance on numerous downstream tasks, with little or no task-specific tuning \citep{Petroni2019LanguageMA,Brown2020LanguageMA}.

There are a number of hypotheses explaining the power of pretrained LLMs. For example, one can postulate that the pretraining data is huge and the model \emph{might} be shallowly memorizing patterns in data \citep{Bender2021OnTD,Carlini2021ExtractingTD}. Alternatively, the LLMs \emph{might} be learning to reason through observed patterns in the pretraining data in novel ways \citep{McCoy2021HowMD}. However, what exact pattern is memorized or reasoned through, especially in an arbitrary downstream task, is not very clear---the \emph{evidence} of these conjectures remains underexplored. 
Such evidence is useful as it may facilitate the trustworthiness of the models \citep{lipton2018mythos}. Moreover, it can surface problematic patterns in data or model behavior and can aid researchers to improve the model \citep{zhong2019fine,Han2021InfluenceTD}.

In this work, we develop a methodology to provide such evidence. 
Our hypothesis is that among the enormous pretraining corpora, there is a subset of pretraining data that influences the model's behavior on a downstream task more than the rest of the pretraining data. 
Therefore, our task is to locate a task-specific evidence set---a very small amount of pretraining data that 
particularly impacts the model's performance on the task. 
We can interpret a model by inspecting such evidence sets---whether they contain task-relevant patterns compared to the rest of the corpora.

This approach to model interpretability is different from and complementary to prior research that locates important tokens or spans in the model inputs  \citep{Ribeiro2016WhySI,lundberg2017unified}. 
Sometimes information relevant to explaining the model's decision 
may not be present in the inference-time input. For example, in zero-shot open-domain question answering \citep{Petroni2019LanguageMA}, the answer to a question is not within the input text.

A related line of research focuses on instance attribution \citep{Koh2017UnderstandingBP,Yeh2018RepresenterPS,Pruthi2020EstimatingTD,Han2020ExplainingBB}, where the goal is to find which training examples are most influential to the model's decision on a single test example. 
However, in this work we are interested in locating pretraining data influencing the \emph{whole task} (a test set). 
We seek such ``global'' evidence for the task because given the scale of the pretraining and task data, it could be inefficient or even infeasible to find and inspect the evidence for each of the task examples.\footnote{Directly applying instance attribution methods to the task level may also yield negative results \citep{IFDNS}.
}

We first formulate our problem of finding evidence from pretraining data by defining an incremental impact of the evidence (\Sref{sec:formulation}). 
We propose a novel method \methodname{}\footnote{Named after the marine mammal for n\textbf{O} pa\textbf{R}ti\textbf{C}ular re\textbf{A}son.} that effectively identifies the evidence by iteratively using task-specific gradient information (\Sref{sec:method}). 
We focus on two classification tasks, sentiment analysis and textual entailment, in a prompt-based setup (\Sref{sec:experimental_setup}). 
We show the effectiveness of the evidence set discovered by \methodname{}, compared with random data subsets and nearest neighboring data in an embedding space (\Sref{sec:evaluation}). 
Our analyses into the discovered evidence show that our experimented model BERT \citep{Devlin2019BERTPO} has an interestingly high reliance on one of its two pretraining corpora \citep[BookCorpus,][]{Zhu_2015_ICCV} as well as on pretraining examples that mask out synonyms to the task verbalizers \citep{Schick2021ExploitingCF} (\Sref{sec:analysis}).

\section{Problem Formulation}
\label{sec:formulation}

Large pretrained language models have been performing increasingly well in a collection of downstream tasks under few-shot or even zero-shot setups \citep{Petroni2019LanguageMA,Brown2020LanguageMA}. Void of the conventional finetuning, they must have directly learned some useful knowledge from the pretraining. However, the pretraining data is seldom curated, instead sourced from mixed domains and prone to noise. It remains a gap to identify what exact pretraining data (if any) lead to a model's competence in a specific downstream task. Finding such pretraining data for a given model and task, which we call \emph{supporting data evidence}, is the objective of this work.

Let us assume we have a pretrained language model $\theta^{\text{PT}}$ that uses a pretraining dataset $D^{\text{PT}} \ni (x_{\text{context}}^{\text{PT}}, x_{\text{masked}}^{\text{PT}})$. For example, for an autoencoding language model \citep[e.g.,][]{Devlin2019BERTPO}, $x_{\text{context}}^{\text{PT}}$ can be a block of text with certain tokens masked, and $x_{\text{masked}}^{\text{PT}}$ can be those tokens in their original forms, waiting to be reconstructed. 
For an autoregressive language model \citep[e.g.,][]{radford2018improving}, $x_{\text{context}}^{\text{PT}}$ can be a preceding observed context, and $x_{\text{masked}}^{\text{PT}}$ can be the next token to be predicted. 
A language model $\theta^{\text{PT}}$ is trained to minimize a loss $\mathcal{L}$ over the pretraining examples, $\theta^{\text{PT}} = \argmin_{\theta} \mathcal{L}(D^{\text{PT}}; \theta)$.

The language model can be applied to many downstream tasks without finetuning any specific modules, via prompting \citep{Schick2021ExploitingCF,Liu2021PretrainPA}. Let us assume we have a dataset of a downstream task for evaluation purposes, $D^{\text{task}} \ni (x^{\text{task}}, y^{\text{task}})$. The language model could make decisions for the task by measuring $p_{\theta}(\text{verbalizer}(y^{\text{task}}) \mid \text{template}(x^{\text{task}}))$. The template supplies a prompt tailored to the task for the model, and the verbalizer maps the output of the language model to the task's label space.\footnote{Our tasks are classification problems in this work, but the framework should be extendable to generation problems as well.}

Our goal is to find the \emph{supporting data evidence} $S$ for the task $D^{\text{task}}$. This evidence should be a set of examples within the pretraining data ($S \subset D^{\text{PT}}$), and we want the size of the set to be very small to give us clearer signals to interpret ($|S| \ll |D^{\text{PT}}|$). 
The supporting data evidence $S$ should ``contribute'' significantly to the performance of the model on the downstream task.

However, we first observe that defining this contribution is already a non-trivial problem. Prior work in instance attribution like influence functions \citep{Koh2017UnderstandingBP} often adopt a ``leave-one-out'' perspective \citep{cook1977detection}. In our case this would mean removing $S$ from $D^{\text{PT}}$, retraining a new language model from scratch, and testing it on $D^{\text{task}}$. 
This is prohibitively expensive.\footnote{Moreover, the definition of influence functions and even leave-one-out can sometimes be arguable, especially in deep non-convex models \citep{basu2020influence,K2021RevisitingMF}.}

In this work, we adopt an ``upweight'' perspective. We want to upweight certain pretraining examples (e.g., $S$) by letting the model see them more times (instead of less times in leave-one-out). 
To avoid retraining the whole language model, we append the selected pretraining examples $S$ to the end of the existing pretraining procedure, letting the model see and update upon them \emph{one} more time. We want to keep such additional updates as minimal as possible to prevent overfitting. 
Specifically, if we randomly batch the supporting data evidence $S$ to mini-batches, we can get a ``boosted'' model via a very small number of optimizer updates: 
\begin{align*}
    \theta_{\text{boosted}}^{\text{PT}} \gets \theta^{\text{PT}} + \textit{updates}_{\theta, \mathcal{L}}(\textit{batched}(S))
\end{align*}
For the simplicity of notation, we fold a sequence of optimizer updates into $\textit{updates}_{\theta, \mathcal{L}}$ that depends on a sequence of data $\textit{batched}(S)$, the model parameters $\theta$, the loss function $\mathcal{L}$, and an optimization algorithm used during pretraining.

The quality of the data evidence $Q(S)$ is the performance difference in the downstream task between the boosted model and the original pretrained model, measured by the original metric associated to the task: 
\begin{align*}
    Q(S) = \text{metric}(D^{\text{task}}; \theta_{\text{boosted}}^{\text{PT}}) - \text{metric}(D^{\text{task}}; \theta^{\text{PT}})
\end{align*}

\paragraph{Comparison}
Our formulation may seem similar to prior work in task-enhancing pretraining \citep{Han2019UnsupervisedDA,Gururangan2020DontSP,Yao2021NLPFS}. However, such methods typically use a large amount of loosely relevant pretraining data (in addition to the in-task training data), for a performance purpose. We instead want to find an orders-of-magnitude-smaller set of pretraining data, providing a clearer signal of their impact on the task for interpretability purposes. 

\paragraph{Implication}
Does our formulation imply that the data outside our defined $S$ have no contributions to the model on the task? Not necessarily. For instance, there could be certain examples that exceptionally help the model capture the grammar of the language, an indirect contribution to the task. However, they are rather unlikely to be picked as supporting data evidence: in the scope of this paper, we focus on finding the evidence that is \emph{directly} related to the downstream task. The indirect evidence is also interesting and may be valuable for future work to investigate further. 

\paragraph{Alternatives}

One issue with our problem formulation is that the \emph{history} of the model pretraining is ignored. 
Is it sufficient to check whether the evidence data help the \emph{fully} pretrained model? What if a data is truly related to the task but got overfitted during earlier stages of pretraining, so continuing pretraining on it would not change the model? 
We think this is possible, but only to some extent and with a limited risk, since 
the huge volume of pretraining data can make it difficult to overfit to all examples related to an arbitrary task in a particular way. 
Nevertheless, to improve that case, a potentially better solution would be ``boosting'' \emph{each checkpoint} of the model across the whole pretraining procedure. This would share an intuition with some instance attribution methods based on model checkpoints \citep{Pruthi2020EstimatingTD}. 
However, we lack the resources to perform such experiments in this work, so we defer that to future research.\footnote{With unconstrained resources, one can extend the checkpoint proposal and even measure Shapley values \citep{shapley1951notes} of the data, an equitable valuation method \citep{ghorbani2019data}.} 

\medskip

With this problem formulation, we are now facing the challenge of actually finding the supporting data evidence $S$ that contribute most to the model's downstream task performance. This is again non-trivial since there are ${|D^{\text{PT}}| \choose |S|}$ possible combinations for $S$, and the verification process introduced above is also not cheap. 
We now propose a novel method \methodname{} identifying the supporting data evidence $S$.

\section{\methodname{}~\includegraphics[scale=0.09]{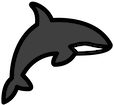}}
\label{sec:method}

\begin{figure}[t]
    \centering
    \includegraphics[width=0.42\textwidth]{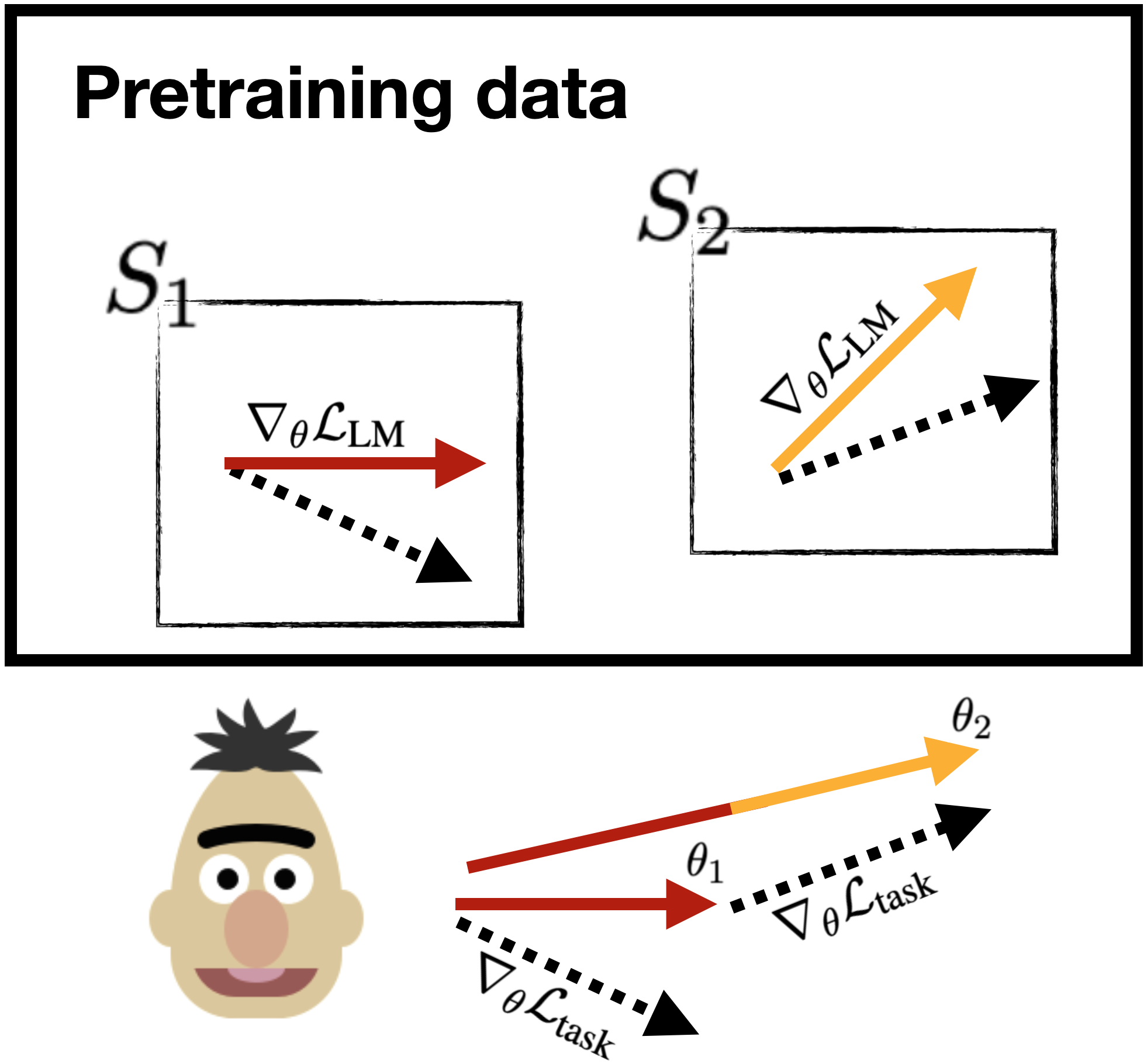}
    \caption{The first two iterations of our proposed method \methodname{}, finding the pretraining data exerting similar gradients to the model as the task data. At each iteration $i$, we continue pretraining the original language model by $\cup_{j=1}^{i}S_{j}$. 
    }
    \label{fig:our_method}
\end{figure}

The goal of our method is to find a very small subset of the pretraining data that is directly helpful to the downstream task when we continue pretraining the language model over it. 
In other words, $S \subset D^{\text{PT}}$, $|S| \ll |D^{\text{PT}}|$, and $Q(S)$ is high. The intuition behind our method is simple:
\begin{enumerate}
    \item We want to find examples in $D^{\text{PT}}$ that exert a similar change to the model parameters as $D^{\text{task}}$. A hypothetical continued pretraining using $D^{\text{task}}$ would likely improve the model on the task and can thus be a good guide.
    \item There could be multiple types of pretraining data that, in conjunction, are useful to the task. We may want to select $S$ in several iterations rather than at once, as a potential continued pretraining with $D^{\text{task}}$ would also be iterative (gradient descent).
\end{enumerate}

Suppose we want to build our supporting data evidence $S$ in $m$ iterations, $S_1, S_2, \ldots , S_m$; the size of the subset at each iteration is $\frac{|S|}{m}$. Below, we first find examples for the first evidence subset $S_1$. We know that continuing pretraining on the task data $D^{\text{task}}$ directly would likely improve the original model $\theta^{\text{PT}}$ on the task. If we put all task data into a batch and calculate a batch gradient $\nabla_{\theta}\mathcal{L}_{\text{task}}(D^{\text{task}}; \theta^{\text{PT}})$,\footnote{The task loss over a single task example can be $\mathcal{L}_{\text{task}}(x^{\text{task}}, y^{\text{task}}) = -\log p_{\theta}(\text{verbalizer}(y^{\text{task}}) \mid \text{template}(x^{\text{task}}))$.} descending along the gradient direction should improve $\theta^{\text{PT}}$. 
Therefore, here we aim to find a subset $S_1$ of the pretraining data that can exert a similar gradient of the model as $\nabla_{\theta}\mathcal{L}_{\text{task}}(D^{\text{task}}; \theta^{\text{PT}})$: 
\begin{align*}
    S_1 = \{ & d \in D^{\text{PT}} \mid \cos (\nabla_\theta \mathcal{L}_{\text{LM}}(d, \theta^{\text{PT}}), \\
    & \nabla_{\theta}\mathcal{L}_{\text{task}}(D^{\text{task}}; \theta^{\text{PT}}) ) > \delta_1\}
\end{align*}
We measure a cosine similarity between gradients and use $\delta_1$ as a dynamic threshold for selecting $|S_1|$ elements.\footnote{The language model loss over a single pretraining example can be $\mathcal{L}_{\text{LM}}(x_{\text{context}}^{\text{PT}}, x_{\text{masked}}^{\text{PT}}) = -\log p_{\theta}(x_{\text{masked}}^{\text{PT}} \mid x_{\text{context}}^{\text{PT}})$.}

Now with the first data evidence subset $S_1$, we can continue pretraining an intermediate model $\theta_{1}^{\text{PT}}$:
\begin{align*}
    \theta_{1}^{\text{PT}} \gets \theta^{\text{PT}} + \textit{updates}_{\theta}(\textit{batched}(S_1))
\end{align*}

The procedure to find the rest of the data evidence subset $S_i$ with $i=2,3,\ldots,m$ is similar to the above but with one difference: these subsets should be beneficial to the model in a way that is not already captured by $S_1$. 

We know that the intermediate model $\theta_{1}^{\text{PT}}$ should capture information about $S_1$. Therefore, for later iterations we calculate a task batch gradient based on the previous intermediate model, $\nabla_{\theta}\mathcal{L}_{\text{task}}(D^{\text{task}}; \theta_{i-1}^{\text{PT}})$. The data evidence subset at each iteration should again exert a similar gradient:
\begin{align*}
    S_i = \{ & d \in D^{\text{PT}} \mid \cos(\nabla_\theta \mathcal{L}_{\text{LM}}(d, \theta_{\lfloor i-1 \rfloor}^{\text{PT}}), \\
    & \nabla_{\theta}\mathcal{L}_{\text{task}}(D^{\text{task}}; \theta_{i-1}^{\text{PT}}) ) > \delta_i\}
\end{align*}
with $\delta_i$ as a dynamic threshold for selecting $|S_i|$ elements. The $\lfloor i-1 \rfloor$ is a design choice that can allow for a ``lagged'' model (i.e., are we computing the gradient of the LM loss w.r.t. the immediate previous intermediate model or the model several iterations before?). 
This lagging may be helpful to the method's stability. 
For the experiments in this work, $\theta_{\lfloor i-1 \rfloor}^{\text{PT}}$ is by default $\theta^{\text{PT}}$, for a maximum lagging; $\theta_{\lfloor i-1 \rfloor}^{\text{PT}}$ is $\theta_{i-1}^{\text{PT}}$ in cases denoted by \emph{NL} (no lagging). 

At each iteration, having a total of $i$ data evidence subsets, we continue pretraining an intermediate model $\theta_{i}^{\text{PT}}$:
\begin{align*}
    \theta_{i}^{\text{PT}} \gets \theta^{\text{PT}} + \textit{updates}_{\theta}(\textit{batched}(\cup_{j=1}^{i}S_{j}))
\end{align*}
It is worth noting that for every iteration, we continue pretraining over the \emph{original} language model, and the data evidence subsets are unordered.

After the $m$-th iteration, we complete building our full supporting data evidence $S = \cup_{j=1}^{m}S_{j}$. The resulting intermediate model $\theta_{m}^{\text{PT}}$ is essentially the ``boosted'' model of our interest, i.e., $\theta_{\text{boosted}}^{\text{PT}}$ introduced in \Sref{sec:formulation}. 
\autoref{fig:our_method} summarizes our method.

\paragraph{Limitation}
Is \methodname{} guaranteed to find the global optimal supporting data evidence out of the ${|D^{\text{PT}}| \choose |S|}$ candidates? No. However, it can still be useful: we will show our method's comparative advantage on $Q(S)$ over some baseline methods in \Sref{sec:evaluation}.

\section{Experimental Setup}
\label{sec:experimental_setup}

\subsection{Baseline methods}
\paragraph{Random sampling}
We simply sample at random $|S|$ examples from $D^{\text{PT}}$ as the supporting data evidence.

\paragraph{Embedding nearest neighbors}
Enhancing language models using examples with nearest neighboring embeddings is a common approach in domain adaptation of LMs and kNN-LMs \citep{Gururangan2020DontSP,Khandelwal2020GeneralizationTM}. 
We want to find nearest neighboring pretraining examples to the task examples. We define a similarity score as below:
\begin{align*}
    \cos (h_{\text{masked}}(\hat{x}_{\text{context}}^{\text{PT}}), h_{\text{verbalizer}}(\text{template}(\hat{x}^{\text{task}})))
\end{align*}
\begin{itemize}
    \item $h_{\text{masked}}$ is the last hidden representation at the position of the masked pretraining token. 
    \item $h_{\text{verbalizer}}$ is the last hidden representation at the position of the task verbalizer token. 
    \item $\hat{x}_{\text{context}}^{\text{PT}}$ is the pretraining input to the model but containing the ground truth masked token. 
    \item $\text{template}(\hat{x}^{\text{task}})$ is the templated task input but supplying the ground truth verbalized label. 
\end{itemize}
We use the ground truth information here for a fair comparison with our method \methodname{}, where the calculation of gradients involves the ground truth information as well.

Practically, since $|D^{\text{task}}|$ can be large and well over $|S|$, we first sample $t$ examples from $D^{\text{task}}$. Then, for each of the $t$ sampled task examples, we find the top-$k$ nearest neighboring pretraining examples in $D^{\text{PT}}$. Finally, from the pool of the $t \cdot k$ pretraining examples, we sample $|S|$ of them as the supporting data evidence. We additionally have a hyperparameter max-$r$ controlling the maximum allowed data repetitions in the selected data evidence.

\paragraph{\methodname{} with embeddings}
We use the gradient information $\cos (\nabla_\theta \mathcal{L}_{\text{LM}}(.), \nabla_\theta \mathcal{L}_{\text{task}}(.))$ when collecting the supporting data evidence subsets in \methodname{}. We want to know whether we can substitute the gradients with hidden representations. Reusing the notations in the embedding nearest neighbors baseline, we use $\cos (h_{\text{masked}}(.), \frac{1}{|D^{\text{task}}|} \sum_{x^{\text{task}}} h_{\text{verbalizer}}(.))$ for all the gradient cosine operations in \methodname{}. Note that we use the average embeddings of all task examples to replace the batch gradient over all task examples. Our other design decisions of \methodname{} remain unchanged.

\subsection{Language model and downstream tasks}
\paragraph{BERT}
In this work, the language model $\theta^{\text{PT}}$ we use is BERT-large \citep{Devlin2019BERTPO}. We choose it primarily due to the limited computing resources we have---BERT is small both in terms of the number of model parameters and the size of the original pretraining data. Our problem formulation and method are extendable to other language models as well. 

\paragraph{IMDB}
We primarily experiment with two text classification tasks, sentiment analysis and textual entailment. For sentiment analysis, we use the IMDB movie review dataset \citep{maas-EtAl}. The task data $D^{\text{task}}$ here is the IMDB test split containing 25,000 examples. The template for the IMDB examples is ``\emph{It was \texttt{[MASK]}. <REVIEW>}''. The verbalizer maps the reconstruction of the \texttt{[MASK]} token to the label space, $\{ \text{``good''} \rightarrow positive, \text{``bad''} \rightarrow negative \}$. 

\paragraph{MNLI}
For the textual entailment task, we use the MNLI dataset \citep{williams2018}. The task data $D^{\text{task}}$ here is the MNLI matched validation split containing 9,815 examples. The template for the MNLI examples is ``\emph{<PREMISE> \texttt{[MASK]}, <HYPOTHESIS>}''. The verbalizer maps the reconstruction of the \texttt{[MASK]} token to the label space, $\{ \text{``yes''} \rightarrow entailment, \text{``no''} \rightarrow contradiction, \text{``maybe''} \rightarrow neutral \}$. We use the OpenPrompt library \citep{Ding2022OpenPromptAO} to prompt the BERT model with the templates and verbalizers inherited from \citet{Gao2021MakingPL}. 

\paragraph{Zero-shot transfer and prompt tuning}
When we formulate our problem, we are interested in the evidence in \emph{pretraining} that directly impact the pretrained model's performance on the downstream task---a zero-shot transfer scenario. There is no notion of \emph{finetuning} with the in-task training data. However, research in prompt tuning \citep[e.g.,][]{Lester2021ThePO} sometimes folds the usage of in-task training data into the template for the task. They add a sequence of soft embeddings to the beginning of the template and train the embeddings with the in-task training data. The language model in this case is mostly untouched, except that the first input word embedding layer is bypassed. 
Apart from our main experiments with the zero-shot transfer model, as additional experiments we consider such prompt tuning scenarios, finding pretraining data evidence useful for the task when the template is enhanced with some in-task training data.\footnote{Depending on the zero-shot performance of the model on the task, we use different amounts of in-task training data for prompt tuning. For IMDB, we use 100 examples per class, whereas for MNLI, we use 10,000 examples per class.} 

\subsection{Pretraining data} 
\paragraph{Source}
BERT uses the English \emph{Wikipedia} and \emph{BookCorpus} \citep{Zhu_2015_ICCV} as its pretraining data. During pretraining, 15\% of the tokens are randomly masked out to be reconstructed. Though BERT's pretraining data is already small compared to those of many other language models \citep[e.g.,][]{Raffel2020ExploringTL,Gao2021ThePA}, we unfortunately still do not have the resource to process the full dataset. In fact, in this work we only randomly sample 0.5\% of the full pretraining data. 

\paragraph{Format}
During pretraining, BERT would reconstruct the masked 15\% tokens in a sequence in parallel (i.e., the reconstruction loss for each token is independent). From the training perspective, this is efficient. However, this work aims to find the supporting data evidence. We particularly want to know learning the reconstruction of \emph{which} token could most impact the downstream task performance. Therefore, we expand each pretraining data and treat each masked token as a standalone example. More specifically in our setup, $D^{\text{PT}} \ni (x_{\text{context}}^{\text{PT}}, x_{\text{masked}}^{\text{PT}})$. $x_{\text{context}}^{\text{PT}}$ is a sequence of 512 tokens, and $x_{\text{masked}}^{\text{PT}}$ is a single masked token in the sequence. Together this makes $|D^{\text{PT}}| =$ 3,924,635 (with 52,640 unique $x_{\text{context}}^{\text{PT}}$ sequences). We choose at most 2,000 instances from $D^{\text{PT}}$ as the supporting data evidence $S$. 

\subsection{Hyperparameters}
\methodname{} finds $S$ in iterations. In this work we use $m$=20 iterations, with each iteration finding 100 examples from $D^{\text{PT}}$ ($|S|$=2000).\footnote{Since we only work with 0.5\% of the pretraining data, in our actual experiments we do replacement across the 20 iterations (i.e., an example at maximum could appear 20 times).} 

For the embedding kNN baseline, we sample $t$=1000 task examples and choose $k$=\{10, 20, 50, 100\} most similar pretraining data. Within the $t \cdot k$ candidate pool, we sample $|S|$=2000 examples, with a max number of repetitions $r$=\{1, 20, 2000\}. 

During the continued pretraining for all methods, we use a batch size of 16, resulting in at most 125 optimizer updates from the original language model. The learning rate is set at one of BERT's default values 2e-5.

\section{Evaluation}
\label{sec:evaluation}
We evaluate the supporting data evidence $S$, identified using ORCA and the baselines, by quantifying the supportiveness of $S$. 
To objectively measure this supportiveness, we measure $Q(S)$ as defined in \Sref{sec:formulation}. 
Note that this section does not focus on whether or not the discovered data evidence is plausible to humans. We will explore what humans can interpret from the actual data evidence in \Sref{sec:analysis}. 

\autoref{tab:main_result} shows our main results: the performance of our zero-shot language model pretrained additionally on $S$, as identified by different methods. 
We first notice that the performance of our original model is lower on MNLI than on IMDB (normalized by the number of classes). 
This could suggest that the entailment task is intrinsically harder for models that have only been trained on pretraining data. 
We observe a moderate performance improvement using the embedding nearest neighbors method and the embedding version of \methodname{}. The best performance is achieved by our proposed method \methodname{}, especially in the task of IMDB by a large margin. 

\autoref{tab:main_result_prompt_tuning} shows some additional results on the effect of the supporting data evidence $S$ on a prompt-tuned model. 
These results show that, compared to the zero-shot model, a prompt-tuned model is more difficult to improve since the prompt may already be highly specialized towards the task, using the in-task training data. 
The \emph{additional} signals in the pretraining data that are useful to the task can be scarce. 
That said, the pretraining data $S$ identified by \methodname{} still improves the model on IMDB. 

\begin{table}[t]
    \centering
    \begin{tabular}{@{}lp{0.5in}p{0.5in}@{}}
    \toprule
     On zero-shot model & IMDB & MNLI\\
      \midrule
      \textit{\textbf{Null}} & \underline{\textit{73.50}} & \underline{\textit{43.70}} \\
      [10pt]
      \textbf{Random} & 71.25 & 42.56 \\
      [4pt]
      \textbf{Embedding kNN} & 76.55 & 45.15 \\
      [4pt]
      \textbf{\methodname{} w/embeddings} & 75.11 & 43.74 \\
      [10pt]
      \textbf{\methodname{}} \small{(NL)} & \textbf{84.51} & 45.46 \\
      [1pt]
      \small{$0 < |S| \leq 500$} & \small{79.81} & \small{44.85} \\
      [1pt]
      \small{$500 < |S| \leq 1000$} & \small{83.87} & \small{45.64} \\
      [1pt]
      \small{$1000 < |S| \leq 1500$} & \small{84.40} & \small{46.10} \\
      [1pt]
      \small{$1500 < |S| \leq 2000$} & \small{85.17} & \small{46.49} \\
      [6pt]
      \textbf{\methodname{}} & 84.33 & \textbf{46.06} \\
      [1pt]
      \small{$0 < |S| \leq 500$} & \small{81.60} & \small{45.99} \\
      [1pt]
      \small{$500 < |S| \leq 1000$} & \small{83.23} & \small{45.75} \\
      [1pt]
      \small{$1000 < |S| \leq 1500$} & \small{84.42} & \small{46.40} \\
      [1pt]
      \small{$1500 < |S| \leq 2000$} & \small{85.15} & \small{46.26} \\
      \bottomrule
    \end{tabular}
    \caption{Main results (accuracy) of \methodname{} and baselines on the zero-shot model. Numbers in regular fonts are averaged from 5 random seeds, while numbers in small fonts show a trajectory of performance with one seed.}
    \label{tab:main_result}
\end{table}

\begin{table}[t]
    \centering
    \begin{tabular}{@{}lp{0.5in}p{0.5in}@{}}
    \toprule
     On prompt-tuned model & IMDB & MNLI\\
      \midrule
      \textit{\textbf{Null}} & \underline{\textit{87.83}} & \underline{\textit{\textbf{70.19}}} \\
      [10pt]
      \textbf{Random} & 86.06 & 69.07 \\
      [4pt]
      \textbf{Embedding kNN} & 86.53 & 68.93 \\
      [4pt]
      \textbf{\methodname{} w/embeddings} & 87.80 & 68.45 \\
      [10pt]
      \textbf{\methodname{}} \small{(NL)} & 87.65 & 68.79 \\
      [4pt]
      \textbf{\methodname{}} & \textbf{88.10} & 68.61 \\
      \bottomrule
    \end{tabular}
    \caption{Additional results (accuracy) of \methodname{} and the baselines on the prompt-tuned model. All the numbers are averaged from 5 random seeds.}
    \label{tab:main_result_prompt_tuning}
\end{table}

\section{Analysis}
\label{sec:analysis}
While useful in showing the effectiveness of the supporting data evidence $S$, evaluations in \Sref{sec:evaluation} do not provide us with tangible insights about the model itself. 
In this section, we analyze some properties of $S$, and see whether they reflect humans' expectations for the model. 
We first show a few qualitative examples of the evidence discovered by \methodname{} in \autoref{tab:qual_ex}.

\begin{table*}[t]
\begin{center}
\renewcommand{\arraystretch}{1.4}
\begin{tabular}{p{0.07\textwidth}p{0.85\textwidth}}
    \toprule
    {IMDB} & {\small{... we have to think that were awfully lucky as human beings to have the nice precise system. the sloppy system is probably \underline{\textbf{good}} enough for bacteria. it turns out -- much to geneticists surprise -- that lowly bacteria store genes as whole units (weasel) ...}}\\
    {} & {\small{it was the only place she could afford. her meager earnings didnt provide much in the ways of clean, modern style along with the privacy she required. she felt better if she thought about how \underline{\textbf{bad}} it could be. a year ago, shed lived with her mother. anywhere was better than living with her ...}}\\
    \midrule
    {MNLI} & {\small{... he then cut the cord that bound her hands and legs. are you ok to walk? he asked hoping the answer was \underline{\textbf{yes}}. i think so. but im quite stiff, she said. he helped her up. stretch your legs a little. theyll feel better ...}}\\
    {} & {\small{... there was no way to hide the shock on her face, and she knew he saw it by his sigh. ``do you think yourself less than me?'' ``\underline{\textbf{no}}!'' she absolutely didn't but ... he nodded his head. ``i see. you thought i would think you were less than me.'' she was ashamed. ``i'm sorry.'' ...}}\\
    {} & {\small{... he shook his head, incredulous. in fact, he looked like he was considering throttling me. ``you're just not getting it. maybe that's my fault. \underline{\textbf{maybe}} it's because i don't tell you i love you often enough. baby, you're the only `good' thing that i've ever had ...}}\\
    \bottomrule
\end{tabular} 
\end{center}
\caption{
Examples of the supporting data evidence ($S$) discovered by \methodname{} for IMDB and MNLI. 
The masked token ($x_{\text{masked}}^{\text{PT}}$) in each example is underlined. 
The example evidence for IMDB expresses sentiments, while it is less clear whether the example evidence for MNLI is related to entailment. 
}
\label{tab:qual_ex}
\end{table*}

\begin{figure}[t]
    \centering
    \includegraphics[width=0.48\textwidth]{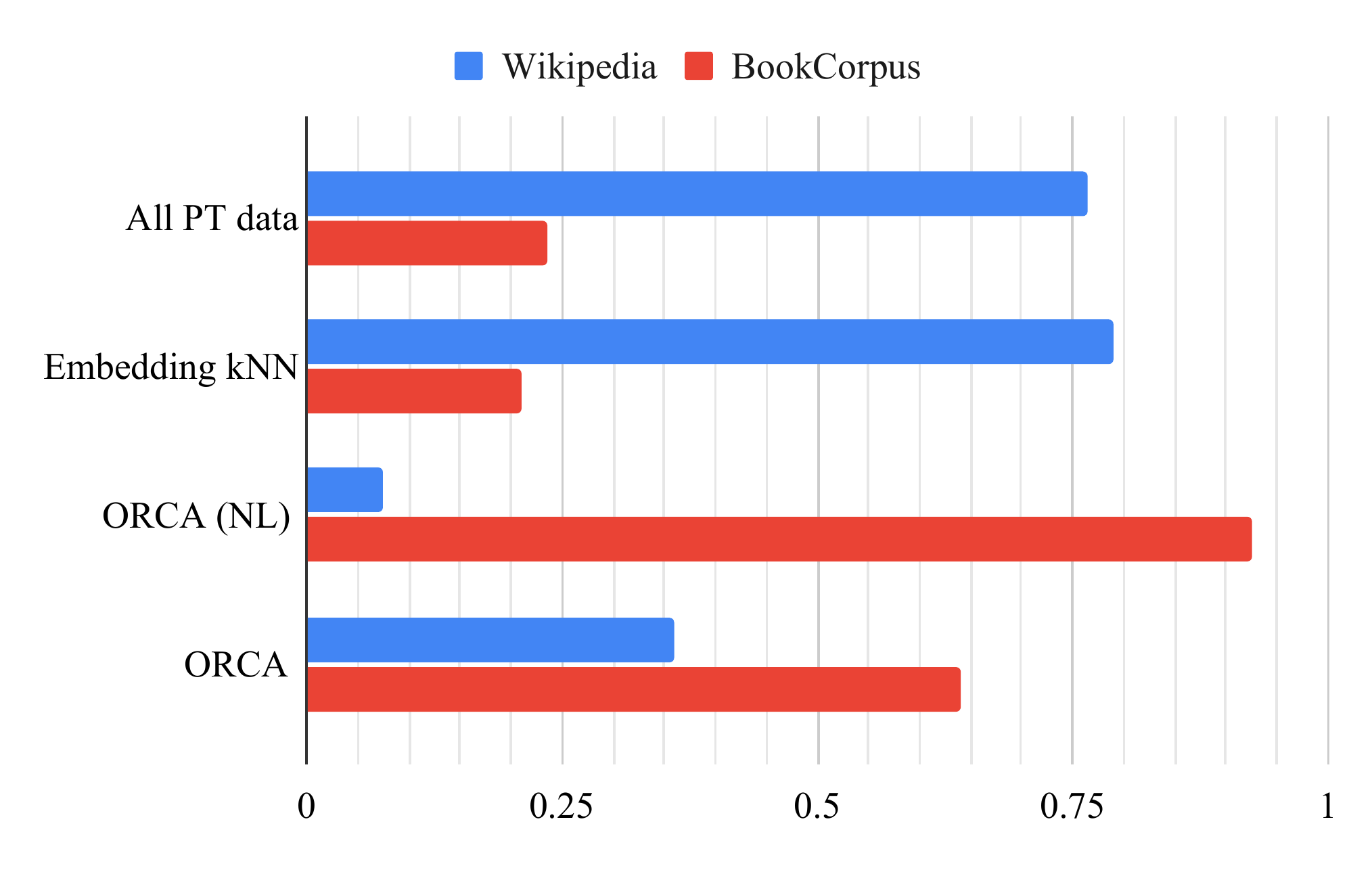}
    \caption{Source corpus distribution of the supporting data evidence in \textbf{IMDB}. 
    }
    \label{fig:source_corpus_imdb}
\end{figure}

\begin{figure}[t]
    \centering
    \includegraphics[width=0.48\textwidth]{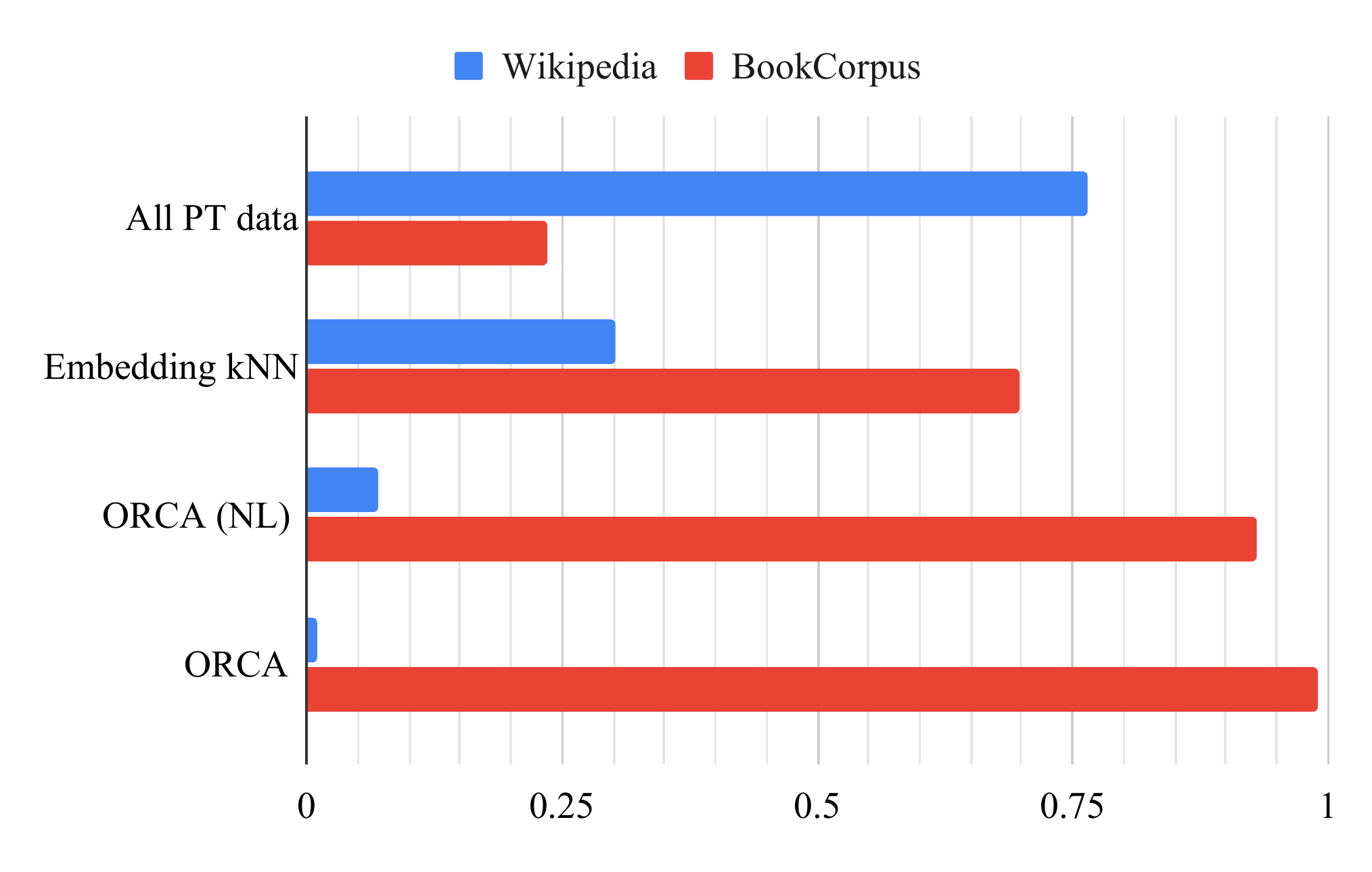}
    \caption{Source corpus distribution of the supporting data evidence in \textbf{MNLI}. 
    }
    \label{fig:source_corpus_mnli}
\end{figure}

\paragraph{Which source corpus does the supporting data evidence come from?} 
The pretraining data of BERT consist of the English Wikipedia and BookCorpus. 
We show the source corpus of examples in $S$ in \autoref{fig:source_corpus_imdb} and \autoref{fig:source_corpus_mnli}. 

We find that though the pretraining set consists of considerably more data from Wikipedia than from BookCorpus (76.5\% vs. 23.5\%), the supporting data evidence identified by \methodname{} has a drastically different source corpus distribution. 
In IMDB, 64.1\% and 92.6\% of the examples in $S$ come from BookCorpus, using the default \methodname{} and its no-lagging variant respectively. 
The demotion of Wikipedia examples in the sentiment analysis task is somewhat reasonable, since Wikipedia is meant to have a neutral point of view (NPOV).\footnote{\url{https://en.wikipedia.org/wiki/Wikipedia:Neutral_point_of_view}} 
On the other hand, BookCorpus consists of novels that could involve strong emotions and sentiments. 

A similar trend can be found in MNLI as well, with 99.0\% and 92.9\% of the examples in $S$ coming from BookCorpus, using the default and NL variant of \methodname{}. 
We conjecture that the over-reliance on BookCorpus in MNLI could be due to the selection of the colloquial verbalizer words (e.g., ``yes'', ``maybe''), which can be scarce in Wikipedia. 
Also, the BookCorpus data could contain more everyday topics that that match MNLI's genres (e.g., fiction, letters, telephone speech). However, whether it is reasonable for the model to rely on BookCorpus for textual entailment is arguable: Wikipedia should be a more reliable source if we want the model to build more upon factual information.

\paragraph{What are the masked tokens in the supporting data evidence?}
Prompted language models use a verbalizer to adapt to the downstream task. For example, outputting ``good'' for a templated IMDB input indicates a positive sentiment, ``yes'' for MNLI indicates entailment, etc. For a pretraining example that supports the task, are there any relations between its masked, to-be-reconstructed pretraining token ($x_{\text{masked}}^{\text{PT}}$) and the verbalizer words for the task ($\text{verbalizer}(y^{\text{task}})$)? 
In \autoref{tab:top_verbalizers_imdb} and \autoref{tab:top_verbalizers_mnli}, we show the 10 most frequent masked words (types) in $S$, for each method in IMDB and MNLI. 

We observe that the verbalizer words, in their original forms, are always the most common masked token in $S$. For all of the methods in both tasks, over 50\% of the masked tokens are exactly the verbalizer words. 
Though we observe some noise in $x_{\text{masked}}^{\text{PT}}$ (e.g., symbols that carry no task-relevant meaning), most of the other masked tokens are synonyms to the verbalizer words in IMDB. 
In MNLI, the other masked tokens may capture relations between clauses similar to the verbalizer words (e.g., then, to, probably). 
Overall, we find that $x_{\text{masked}}^{\text{PT}}$ in the discovered $S$ is reasonable for the sentiment analysis and textual entailment task. 

\begin{table}[t]
\begin{center}
\renewcommand{\arraystretch}{1.3}
\begin{tabular}{p{0.11\textwidth}p{0.31\textwidth}}
    \toprule
    {Method} & {Most frequent $x_{\text{masked}}^{\text{PT}}$ in $S$}\\
    \midrule
    {Embedding kNN} & {\textbf{bad}, \textbf{good}, terrible, great, badly, excellent, worst, negative, better, disappointment, ...
    \small{[\emph{11 distinct tokens in total, 94.8\% \textbf{verbalizer words}}]}}\\
    {\methodname{} \small{(NL)}} & {\textbf{bad}, \textbf{good}, worst, n, worse, ', wrong, -, horrible, poisonous, ...
    \small{[\emph{91 distinct tokens in total, 90.0\% \textbf{verbalizer words}}]}}\\
    {\methodname{}} & {\textbf{bad}, \textbf{good}, \`{}, horrible, not, worse, ugly, hated, poor, terrible, ...
    \small{[\emph{285 distinct tokens in total, 55.9\% \textbf{verbalizer words}}]}}\\
    \bottomrule
\end{tabular} 
\end{center}
\caption{
Masked tokens ($x_{\text{masked}}^{\text{PT}}$) in the supporting data evidence of \textbf{IMDB}.
}
\label{tab:top_verbalizers_imdb}
\end{table}

\begin{table}[t]
\begin{center}
\renewcommand{\arraystretch}{1.3}
\begin{tabular}{p{0.11\textwidth}p{0.31\textwidth}}
    \toprule
    {Method} & {Most frequent $x_{\text{masked}}^{\text{PT}}$ in $S$}\\
    \midrule
    {Embedding kNN} & {\textbf{no}, \textbf{yes}, \textbf{maybe}, \`{}, yeah, However, perhaps, n, ), No, ...
    \small{[\emph{258 distinct tokens in total, 58.3\% \textbf{verbalizer words}}]}}\\
    {\methodname{} \small{(NL)}} & {\textbf{maybe}, \textbf{yes}, \textbf{no}, \`{}, n, that, -, then, perhaps, the, ...
    \small{[\emph{176 distinct tokens in total, 69.1\% \textbf{verbalizer words}}]}}\\
    {\methodname{}} & {\textbf{maybe}, \textbf{yes}, \textbf{no}, \`{}, perhaps, to, probably, has, in, big, ...
    \small{[\emph{125 distinct tokens in total, 59.4\% \textbf{verbalizer words}}]}}\\
    \bottomrule
\end{tabular} 
\end{center}
\caption{
Masked tokens ($x_{\text{masked}}^{\text{PT}}$) in the supporting data evidence of \textbf{MNLI}.
}
\label{tab:top_verbalizers_mnli}
\end{table}

\paragraph{Is the context of the supporting data evidence similar to the task input data?}
We are interested in the relationship between the context of the selected pretraining data ($x_{\text{context}}^{\text{PT}}$) and the input of the downstream task ($x^{\text{task}}$). Are they exceptionally similar, indicating that the model may be memorizing shallow patterns? Alternatively, are they considerably different, indicating that the model needs to transfer some learnt knowledge from pretraining to the task (either in a reasonable or spurious way)? 
Our exploratory step uses an automatic metric between two distributions of texts, MAUVE \citep{Pillutla2021MAUVEMT}, to measure the similarity between our sets of $x_{\text{context}}^{\text{PT}}$ and $x^{\text{task}}$. 
As a method based on quantized language model embeddings, MAUVE similarity may capture text attributes such as topics and style.\footnote{Grammaticality can be another attribute as \citet{Pillutla2021MAUVEMT} work with machine-generated texts. This is less relevant in our case as our sets of texts are naturally occurring.} 

\begin{figure}[t]
    \centering
    \includegraphics[width=0.48\textwidth]{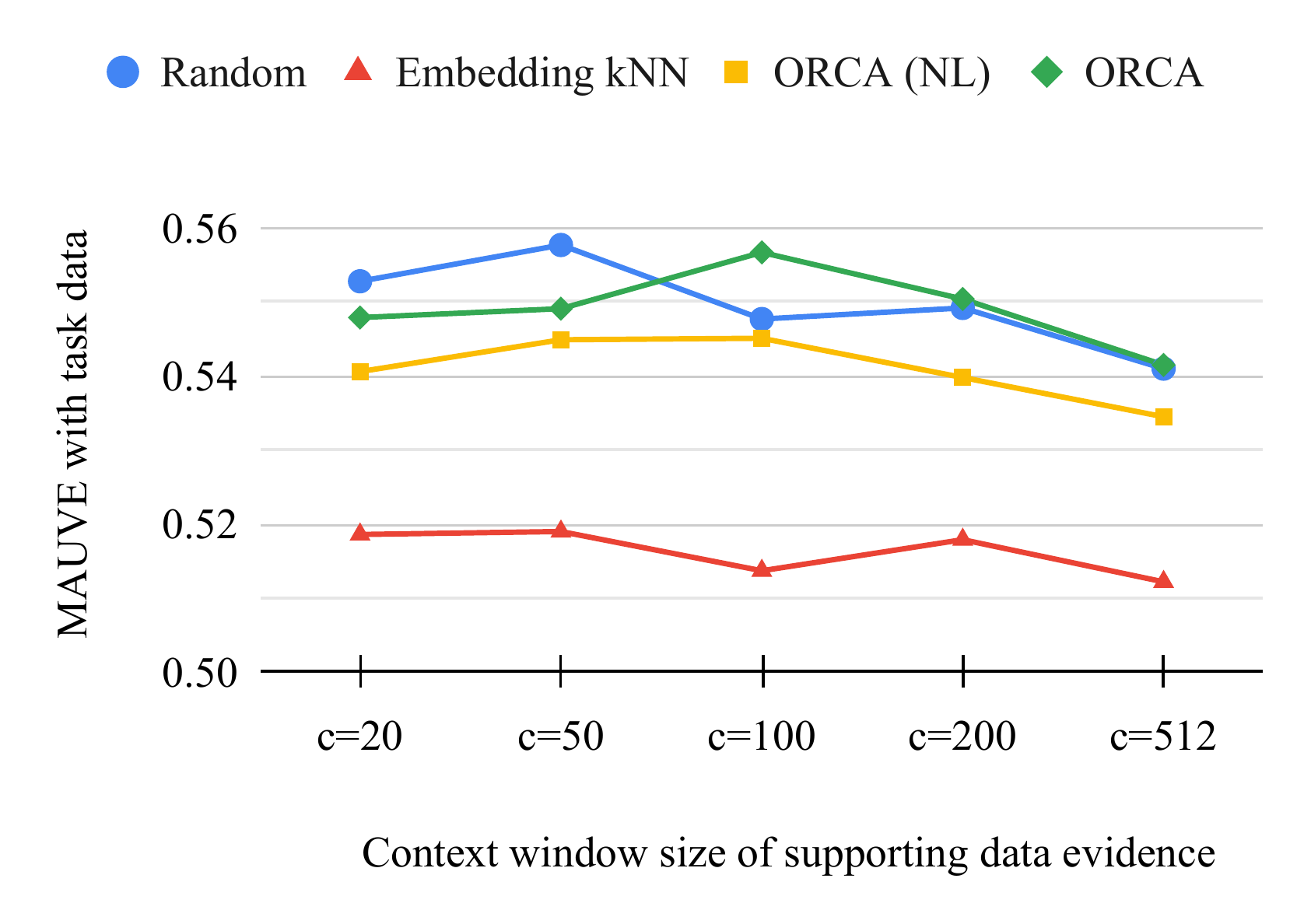}
    \caption{MAUVE similarity on \textbf{IMDB}, between the sets of $x_{\text{context}}^{\text{PT}}$ in $S$ and $x^{\text{task}}$. 
    }
    \label{fig:mauve_imdb}
\end{figure}

\begin{figure}[t]
    \centering
    \includegraphics[width=0.48\textwidth]{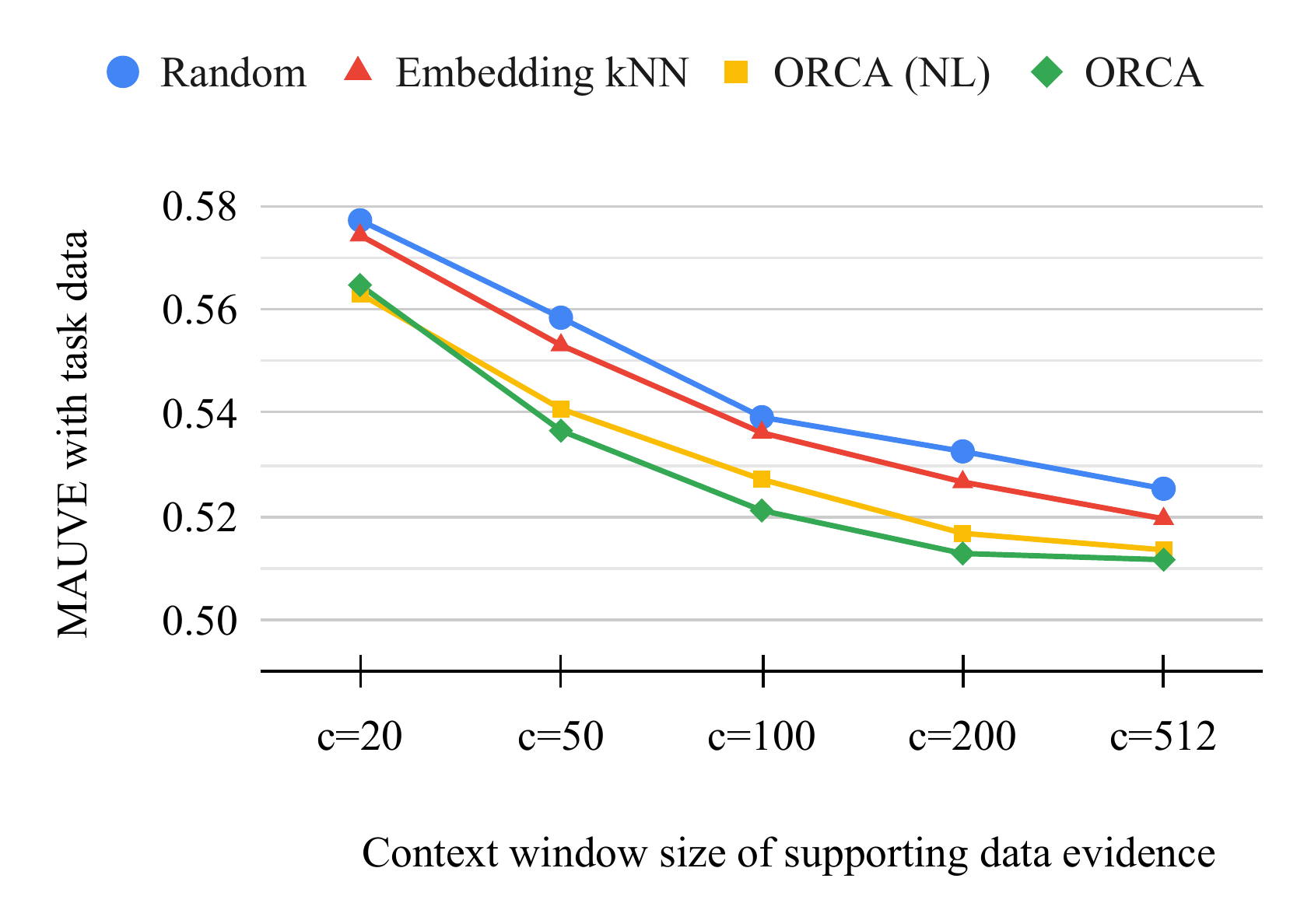}
    \caption{MAUVE similarity on \textbf{MNLI}, between the sets of $x_{\text{context}}^{\text{PT}}$ in $S$ and $x^{\text{task}}$. 
    }
    \label{fig:mauve_mnli}
\end{figure}

Apart from using all 512 tokens in the context of the data evidence ($x_{\text{context}}^{\text{PT}}$), we also truncate the context, keeping the surrounding $c$ tokens of the masked token ($x_{\text{masked}}^{\text{PT}}$). We want to control for the scope of the context by varying $c$. For $x^{\text{task}}$, we randomly sample 2000 examples to match the size of $S$. 
\autoref{fig:mauve_imdb} and \autoref{fig:mauve_mnli} show the results on IMDB and MNLI. 

We observe that the MAUVE scores between $x_{\text{context}}^{\text{PT}}$ and $x^{\text{task}}$ are all between 0.512 and 0.577. In contrast, the MAUVE score between the training set of the task and the test set ($x^{\text{task}}$) is 0.998 and 0.997 for IMDB and MNLI respectively. This substantial difference in MAUVE scores may indicate a disparity in topics and style between the context of the pretraining evidence and the task data. 
Additionally, the MAUVE score of our selected supporting data evidence is not higher than a random sample in most cases. This further shows that the signal in the evidence context useful for the task is \emph{subtle}, in a way that MAUVE cannot capture. While not within the scope of this paper, future investigation can also extend the analysis of the supporting evidence with feature attribution methods \citep{Pezeshkpour2022CombiningFA} or a human evaluation with domain experts of the task.

\section{Related Work}

LLMs have been showing competence in various downstream tasks in NLP with little to no task-specific tuning, using prompts \citep{Petroni2019LanguageMA,Brown2020LanguageMA,Schick2021ExploitingCF,Gao2021MakingPL,Lester2021ThePO}. We are especially interested in interpreting LLMs under a zero-shot setup, where the knowledge relevant to the downstream task must come from the noisy pretraining data.\footnote{Interpreting the role of pretraining data in an unprompted, finetuning setup can be intrinsically harder, but prior work like \citet{chen2020multi} have made attempts.} 

One common interpretability method for NLP models is feature attribution, where important tokens or spans in the inference-time input are highlighted, indicating their contributions to the model's decision \citep{Simonyan2013DeepIC,Li2016VisualizingAU,Ribeiro2016WhySI,lundberg2017unified}. 
Another type of interpretation that aligns more with our focus is instance attribution, where important training examples are highlighted for their influence on the model \citep{Koh2017UnderstandingBP,Yeh2018RepresenterPS,Pruthi2020EstimatingTD,Han2020ExplainingBB,Guo2021FastIFSI}. 
In this work, we are instead interested in the influence of pretraining data and in finding supporting data evidence for the entire task rather than individual test examples.\footnote{A recent concurrent work by \citet{Akyrek2022TracingKI} builds a candidate set for fact-tracing in question answering; the difference is the use of task-related training examples instead of pretraining data, and the evaluation with an information retrieval objective.} 
There has also been prior work analyzing what amount of data is needed during pretraining to achieve models with certain capabilities \citep{Zhang2021WhenDY}, but these works do not attribute model performance to specific pretraining data. 

Our proposed method to find the data evidence, \methodname{}, shares a similar intuition with prior work that reweighs training data \citep{wang2020optimizing}, as both methods use the gradient information of the test data. 
However, their target model depends on an \emph{ordered sequence} of data weights and model checkpoints. In contrast, we apply an \emph{unordered} data evidence set to the \emph{original} model, mimicking an upweighting in pretraining. The root of the difference is the purpose: the former is performance-oriented while the latter is interpretability-oriented. 

Another related line of work in machine learning is coreset construction \citep{Coleman2020SelectionVP,Mirzasoleiman2020CoresetsFD,Huang2021ANS}. Their focus is typically an empirical risk minimization problem on the training data, without a notion of downstream tasks. They aim to create a substitution set for the full training data for an efficiency purpose.

\section{Conclusion}
The competence of zero-shot or few-shot prompted language models on various downstream tasks is mysterious. 
The models should be gaining task-specific knowledge from the pretraining data, 
but what pretraining data leads to the capability of the models is an underexplored area of research. 
In this work, we formulate the problem of finding supporting data evidence in the pretraining data of LLMs for downstream tasks. 
We propose \methodname{} to effectively identify such evidence with an iterative guide from task-specific gradient information. 
Further analyses into the evidence show that a prompted BERT on sentiment analysis and textual entailment relies heavily on the BookCorpus data, as well as on pretraining examples that mask out task verbalizers and their synonyms. 

There remain several gaps to be addressed by future work. For example, the definition of the data evidence quality can be more theoretically grounded (we discussed a few alternatives in \Sref{sec:formulation}). 
Our proposed method \methodname{} is slow and currently operates on a small amount of pretraining data, since it computes a per-sample gradient for each example in the corpus. 
We did not explore many potential design choices of the method (e.g., warmup selections, annealing mechanisms). 
Finally, while our analysis into the identified supporting data evidence successfully captures high-level signals in the pretraining data (e.g., source corpus, masked token, whole context), this can be complemented with more detailed interpretations (e.g., what exact spans in the supporting data evidence contribute to their supportiveness).

\bibliography{my_cites}
\bibliographystyle{acl_natbib}


\appendix



\end{document}